\documentclass[english]{article}
\usepackage{mathptmx}

\usepackage[T1]{fontenc}
\usepackage[latin9]{inputenc}
\usepackage{geometry}
\usepackage{float}
\usepackage{graphicx}
\usepackage{babel}
\begin{document}

\title{\noindent A Bayesian algorithm for detecting identity matches\\
and fraud in image databases%
\thanks{\noindent Copyright Digital Signal Corporation, 14000 Thunderbolt
Place, Chantilly, VA 20151, All rights reserved.%
}}

\author{\noindent Gaurav Thakur%
\thanks{\noindent email: gthakur@alumni.princeton.edu%
}}

\date{\noindent June 1, 2015%
\thanks{\noindent Released in April 2017%
}}
\maketitle
\begin{abstract}
A statistical algorithm for categorizing different types of matches
and fraud in image databases is presented. The approach is based on
a generative model of a graph representing images and connections
between pairs of identities, trained using properties of a matching
algorithm between images.
\end{abstract}

\section{\noindent Introduction}

\noindent This report describes a machine learning algorithm for detecting
and classifying different types of identity fraud, developed as part
of Digital Signal Corporation's DCAF 2.0 product and service (Database
Conversion and the Analysis of Fraud). In this context, detecting
identity fraud amounts to finding different types of connections between
pairs of identities (IDs) across databases of IDs, each containing
various types of images of faces. The algorithm is based on a probabilistic,
generative model of a graph representing all images in a pair of IDs,
coupled with an image matcher, or a program that produces a similarity
score between $0$ and $1$ for any pair of images. The fraud detection
algorithm takes advantage of known statistics of the matcher on a
training dataset and directly computes likelihoods of several types
of fraud that a system operator would be interested in, without producing
or relying on any intermediate feature, which results in a highly
accurate and nearly optimal classifier.\\

\noindent The scenario addressed by the algorithm can be described
as follows. We have a collection of $P$ IDs, the \textit{probe},
to be tested against a (typically much larger) database of $G$ IDs,
the \textit{gallery}. Each ID consists of a collection of 2D images
and/or 3D meshes taken from DSC's laser imaging system, although for
clarity, the results in this report discuss 2D images only. A person
approaches a checkpoint with an ID card containing several existing
images of his or her face, and the system collects one or more additional
images of the face to add to the ID, forming the probe. This may also
be repeated for a sequence of individuals, each having their own ID,
to form a larger probe with $P>1$. The gallery is an existing database
of IDs previously collected by the system in this manner. For each
pair of a probe ID and a gallery ID, the classifier determines if
there is a match, and if so, what the nature of the match is. For
example, the two IDs might actually be the same individual and all
of the images within them would be (approximately) the same, or the
person could have possession of someone else's ID card, so that it
does not match their own face but instead matches another ID in the
gallery database. This matching process is performed on all $PG$
pairs of IDs, and the outputs are a $P\times G$ matrix of decisions
as well as a numerical score for each pair that was identified as
fraud, indicating how strong that result was. For each fraud type
of interest, a ranked list of probe-gallery ID pairs with the top
scores are submitted to an operator for further inspection. The subsequent
sections of this report describe the different types of identity fraud
the system looks for, the mathematical and statistical framework used
to model these classes, and various performance tests and simulations.

\section{\noindent Types of identity fraud}

\noindent Let $R$ be the edges of the observed graph, consisting
of a probe ID and a gallery ID with vertices $V_{1}$ and $V_{2}$
respectively. Let $E_{1}$ and $E_{2}$ be the edges within the probe
and gallery IDs and $E_{1,2}$ be the edges between the two IDs, so
that $R=E_{1}\cup E_{2}\cup E_{1,2}$. The vertices represent images
and the edges are the matcher scores between pairs of images, which
are assumed to be between $0$ and $1$. Note that if $|V_{1}|=N_{P}$
and $|V_{2}|=N_{G}$, where $|\cdot|$ is the size of a set, then
we can calculate the total number of edges in $R$ as $\frac{1}{2}N_{P}(N_{P}-1)+\frac{1}{2}N_{G}(N_{G}-1)+N_{P}N_{G}$.
Given $R$, the classifier is built to distinguish between seven possible
hypotheses $H$, including a baseline ``no fraud'' case and six
different types of fraud.\\

\noindent The ``No fraud'' case, $H=1$, is the situation where
the probe and gallery subgraphs are each complete and also fully disjoint
from each other. This is the baseline case that we would expect to
see when the two IDs are in fact different people. ``Multi-ID,''
or $H=2$, is the case where both subgraphs are complete and fully
connected to each other, i.e. the two IDs are actually the same person.
``Probe mismatch,'' $H=3$, is the case when the probe subgraph
is incomplete, but the two IDs are fully disjoint from each other.
This represents a situation where the probe ID contains fraudulent
images, but the gallery ID has no involvement in the fraud, and we
must look at other gallery IDs to locate the match. ``Probe mixed-ID,''
$H=4$, is similar to $H=3$ but where the vertices that are disjoint
in the probe subgraph have full connections to every image in the
gallery subgraph. This occurs when we have found the gallery ID that
the mismatched images in the probe ID belong to. ``Gallery mismatch,''
$H=5$, and ``Gallery mixed-ID,'' $H=6$, are the same as $H=3$
and $H=4$ respectively but with the roles of the probe and gallery
reversed. Note that the interpretation of $H=5$ can vary in practice.
It is not necessarily a situation we want to call fraud (especially
if we are only interested in finding fraud within the probe ID) but
it has a fairly high probability of occurring in practice, and needs
to be accounted for by the classifier to obtain accurate results with
the other hypotheses. The final case, ``Crossed ID'' or $H=7$,
occurs when both subgraphs are incomplete, but there are two disjoint
linkages between them. It corresponds to a scenario where a group
of people (such as a family) is enrolling under multiple IDs. Examples
of the seven hypotheses are shown in Figure 1 for an ID pair with
two images in each ID.\\
\begin{figure}
\centering{}\includegraphics[trim=0in 1.4in 0in 1.4in, clip=true, scale=0.5]{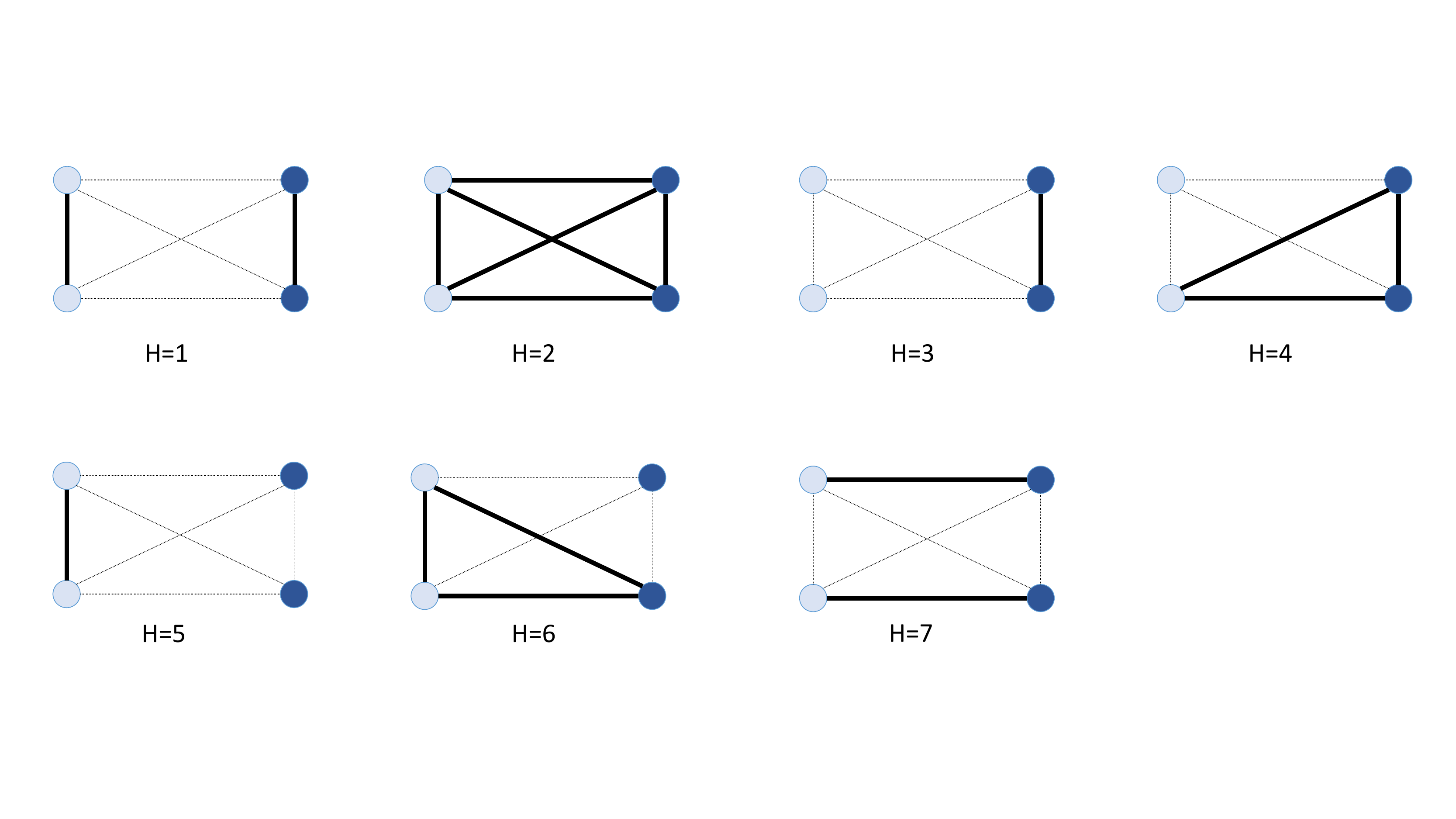}\protect\caption{The seven hypotheses that the classifier chooses between, for a (2,2)
ID pair. The vertices on the left are in the probe ID and the ones
on the right are in the gallery ID.}
\end{figure}

\noindent Note that these hypotheses describe the underlying ground
truth on the graph. Due to the slight inaccuracy and randomness of
the matcher, it is possible for ``nonphysical'' data to be actually
observed in practice, such as a case with three images, A, B and C,
where the edges A-B and B-C have scores of 0.9, but A-C has a score
of 0.1. The classifier will still map this observation to the hypothesis
$H$ that is the best match to the data. It should also be remarked
that the seven hypotheses do not cover every possible ground truth.
For example, one can imagine a case similar to $H=7$ where there
are three or more disjoint linkages. However, in practice we expect
that cases like this have a very low probability of occurring, and
even when they do occur, the classifier will still select the ``closest''
hypothesis, usually $H=7$.\\

\noindent This type of graph is formed for every pair between the
$P$ probe IDs and $G$ gallery IDs, and the resulting $PG$ problems
are largely treated independently of each other. Note that in many
other contexts, similar types of matching problems are typically addressed
by using hash tables, reducing the $PG$ ID comparisons to a number
on the order of $P+G$. However, in the scenario we address here,
the matcher scores and classifier outputs are approximate and real-valued.
Conventional hash functions preserve exact matches between data but
by design, fail to preserve close distances between data, and thus
are not applicable.

\section{\noindent Theory and statistical model}

\noindent We assume we are given $p_{M}$ and $p_{N}$, the match
and non-match sample densities of the image matcher. These statistics
may come from the same probe and/or gallery dataset that we are testing
for fraud in, or from a separate, training gallery dataset. The training
data only consists of the matcher outputs for pairs of matching and
non-matching images, and does not require any information about fraud.
The sample statistics are formed using histograms of the data with
$CN^{1/3}/\sigma$ bins for $N$ samples, where $C$ is some constant
and $\sigma$ is the sample standard deviation; several rules of this
type exist in the statistics literature and have various optimality
properties. Smoother density estimates such as those based on Gaussian
or wavelet kernels were investigated, but turned out to be less well
suited than histograms for capturing the sharp spikes that typically
appear in these densities (see Figure 2).\\

\noindent We now make the assumption that the scores on every edge
are independent, similar to a naive Bayes classifier. This is an approximation
and is not the case in practice, but it turns out to affect all the
hypotheses roughly equally and does not significantly impact the classifier's
accuracy. The likelihood function of $R$ under each hypothesis is
not easy to estimate directly, but this assumption allows us to express
it entirely in terms of the densities $p_{M}$ and $p_{N}$. We define
$V_{1},V_{2},E_{1},E_{2},E_{1,2}$ as in the previous section, with
$R=E_{1}\cup E_{2}\cup E_{1,2}$. For any collection of vertices $V$,
we also denote the reduced power set (collection of all subsets, except
the empty set and the entire set) of $V$ by $2^{V}-2$, and the set
of edges in the complete graph on $V$ by $E(V)$. Note that $|2^{V}-2|=2^{|V|}-2$.
Furthermore, for any collection of edges $E\subset R$, define the
match likelihood on those edges by
\[
L(E)=\left(\prod_{E_{i}\in E}p_{M}(E_{i})\right)\left(\prod_{E_{i}\in R\backslash E}p_{N}(E_{i})\right).
\]

\noindent The classifier first computes the likelihoods $p(R|H=h)$
for each of the seven hypotheses $h$. The likelihoods are given by
the following formulas.

\noindent 
\begin{eqnarray*}
p(R|H=1) & = & L(E_{1}\cup E_{2})\\
p(R|H=2) & = & L(E_{1}\cup E_{2}\cup E_{1,2})\\
p(R|H=3) & = & \frac{1}{|2^{V_{1}}-2|}\sum_{S\in2^{V_{1}}-2}L(E(S)\cup E_{2})\\
p(R|H=4) & = & \frac{1}{|2^{V_{1}}-2|}\sum_{S\in2^{V_{1}}-2}L(E(S)\cup E(V_{1}\backslash S\thinspace\cup\thinspace V_{2}))\\
p(R|H=5) & = & \frac{1}{|2^{V_{2}}-2|}\sum_{S\in2^{V_{2}}-2}L(E(S)\cup E_{1})\\
p(R|H=6) & = & \frac{1}{|2^{V_{2}}-2|}\sum_{S\in2^{V_{2}}-2}L(E(S)\cup E(V_{2}\backslash S\thinspace\cup\thinspace V_{1}))\\
p(R|H=7) & = & \frac{1}{|2^{V_{1}}-2||2^{V_{2}}-2|}\sum_{S_{1}\in2^{V_{1}}-2}\sum_{S_{2}\in2^{V_{2}}-2}L(E(S_{1}\cup V_{2}\backslash S_{2})\cup E(S_{2}\cup V_{1}\backslash S_{1}))
\end{eqnarray*}

\noindent Note that these formulas match the structure of the hypotheses
in Figure 1. For example, under $H=1$, the likelihood function reflects
the fact that we expect $E_{1}$ and $E_{2}$ to all be matches and
$E_{1,2}$ to all be non-matches. The sums run over all possible subsets
of fraudulent images within each ID, that correspond to valid cases
under the given hypothesis. The number of terms grows exponentially
large with the number of images, making the likelihoods potentially
expensive to compute. One way to simplify this problem is to not consider
all elements in $2^{V}-2$, but restrict the sums to subsets containing
at most $M$ elements for some fixed $M$, i.e. only considering cases
where any given ID has at most $M$ fraudulent images. The effect
of this approximation will be investigated in the next section, but
it turns out to be a reasonable assumption in practice, where even
large IDs typically contain at most only a few mismatched images.
In practice, the vertices and indices corresponding to each element
of the reduced power set can be stored in a lookup table and reused
across different pairs of identities, and ordered by the size of each
subset using the ``revolving door algorithm'' based on Pascal's
triangle.\\

\noindent Once the likelihoods are computed, they can be used to make
a decision $D$ about $H$ according to one of several standard optimality
criteria: the maximum likelihood (ML), maximum a posteriori (MAP)
or minimum mean square (MMS) estimates, which correspond to the following
choices for $D$.
\begin{eqnarray*}
D_{\mathrm{ML}} & = & \arg\max_{h}p(R|H=h)\\
D_{\mathrm{MAP}} & = & \arg\max_{h}p(R|H=h)p(H=h)\\
D_{\mathrm{MMS}} & = & \mathrm{round}\bigg(\frac{\sum_{h=1}^{7}hp(R|H=h)p(H=h)}{\sum_{h=1}^{7}p(R|H=h)p(H=h)}\bigg)
\end{eqnarray*}

\noindent Note that $D_{\mathrm{MMS}}$ is effectively an average
and depend on what order the 7 classes are in (i.e. which one is $H=1$,
$H=2$ and so on). $D_{\mathrm{MAP}}$ corresponds to the so-called
naive Bayes classifier, and is the one the DCAF system uses in practice.
If the priors $p(H=h)$ were all equal for $1\leq h\leq7$, then $D_{\mathrm{MAP}}=D_{\mathrm{ML}}$,
but the ``no fraud'' $p(H=1)$ probability is typically much larger
than any of the other cases on common training datasets.\\

\noindent Once the algorithm makes a decision $D$, the log-likelihood
of the corresponding hypothesis $\log p(G|H=D)$ is used as a numerical
score of how strongly fraudulent that pair of IDs is, relative to
other pairs of IDs. These scores are then used to produce a list of
the most strongly fraudulent pairs of IDs. In practice, this log-likelihood
is divided by the total number of edges in the graph to produce a
score that is independent of the number of images in the ID.

\section{\noindent Performance and simulation results}

\noindent This section summarizes some results on the performance
of the classifier under different constraints and datasets. The densities
$p_{M}$ and $p_{N}$ are estimated from the outputs of a commercial
2D image matcher (Cognitec FaceVACS; see \cite{G09} and \cite{C15})
on every pair of faces in the MORPH face dataset (about $80,000$
face images), which produces $136,000$ match and $231$ million non-match
points. Note that the number of non-match points is much larger, as
is typical, and its histogram has more bins to account for this. The
resulting density estimates are shown in Figure 2.

\noindent 
\begin{figure}
\begin{centering}
\includegraphics[scale=0.7]{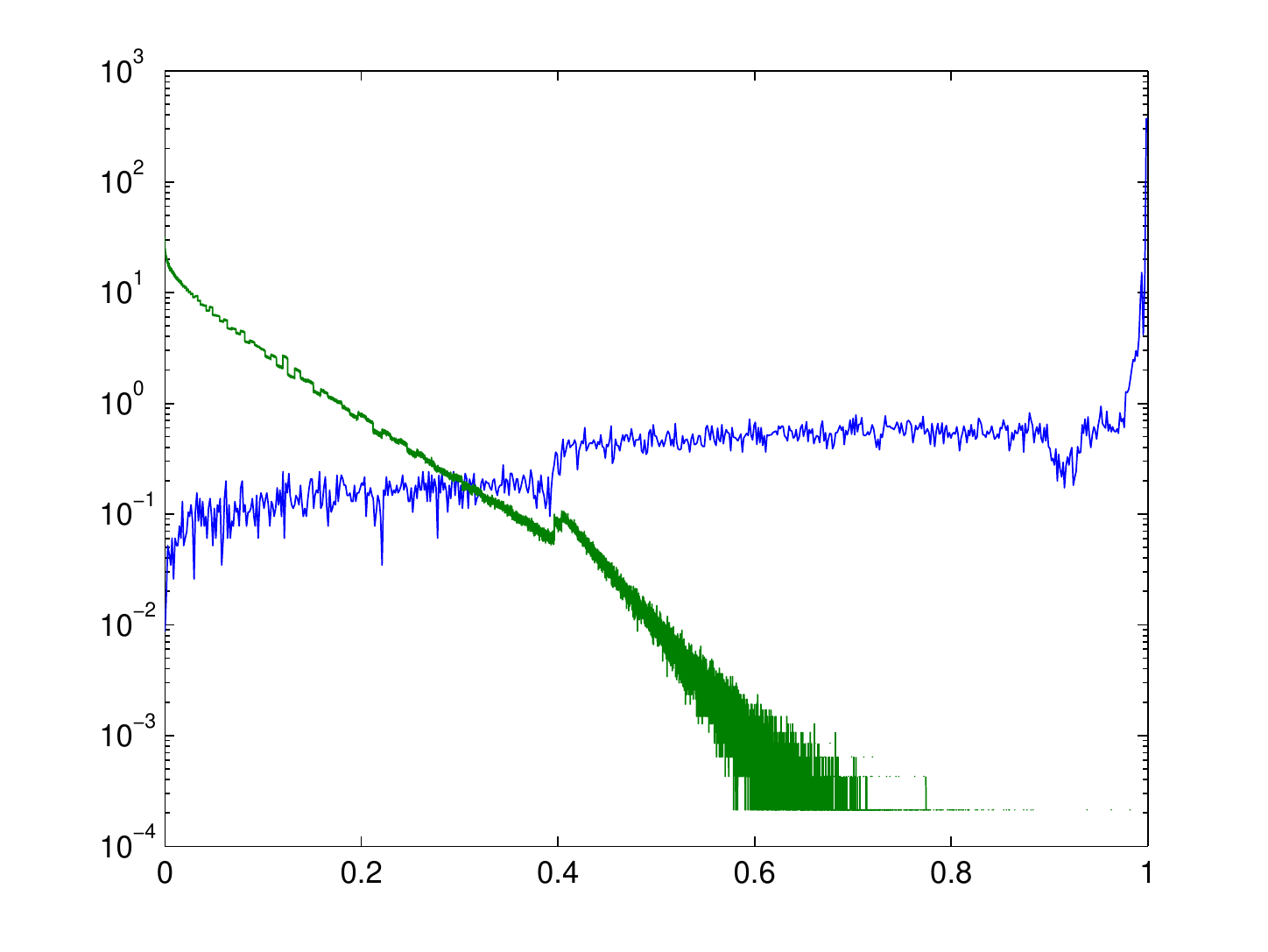}
\par\end{centering}

\protect\caption{Match (blue) and non-match (green) density estimates from MORPH data
under Cognitec.}
\end{figure}

\noindent To test the performance of the classifier, we check the
classifier's decisions on simulated (3,3) ID pairs from several test
datasets, i.e. 3 images each in the probe and gallery IDs. The datasets
consist of a random subset of MORPH (called MORPH\_Rand\_10), a dataset
of the company employees (DSCEmployee) and an older dataset collected
at the University of Notre Dame at 45-degree angle offsets (NDOff45).
The MORPH subset contains 900 identities with 10 images each; we treat
100 of these as the probe and the remaining 800 as the gallery. The
simulation generates fraudulent ID pairs corresponding to 4 of the
7 hypotheses, $H\in\{1,2,4,6\}$. We restrict our attention to only
these fraud cases to allow for a comparison with a legacy algorithm
used in DCAF 1.6 to address the same problem, which uses a spectral
clustering-based feature and only recognizes these cases \cite{R15}.
The maximum likelihood decision $D_{\mathrm{ML}}$ is used for the
statistical classifier, to account for the fact that the simulation
generates an equal number of each of the fraud cases. Note that in
a real system where the fraudulent cases are unlikely to begin with,
$D_{\mathrm{MAP}}$ with non-uniform priors would be expected to have
better performance with lower false positive rates.\\

\noindent The production version of the classifier is implemented
in DCAF 2.0 as a C++ module with Python and Matlab APIs. The likelihoods
for each probe-gallery ID pair in a database can be computed concurrently
and are easily amenable to multithreading or GPU computation. The
inputs are matcher scores between the probe and gallery databases
and the classifier outputs matcher scores. We set the maximum number
of fraudulent images per ID $M=3$, which, as will be seen below,
speeds up the calculations by a few orders of magnitude with a minimal
effect on the classifier's accuracy. The 4x4 matrix of classification
accuracies between different hypotheses $H$ and decisions $D$ is
shown for MORPH\_Rand\_10 and NDOff45 in Figure 3, for both this technique
as well as the clustering method. It can be seen that the statistical
algorithm has significantly better accuracy with both datasets, especially
in terms of the false positives (the top rows in Figure 3) that are
often critical in a real situation. The relative improvement is also
greater on NDOff45, a harder dataset to work with due to its greater
angle variations, indicating that the classifier performs especially
well on borderline cases.

\noindent 
\begin{figure}
\begin{centering}
\includegraphics[trim=3in 2.4in 3in 1.6in, clip=true, scale=1.0]{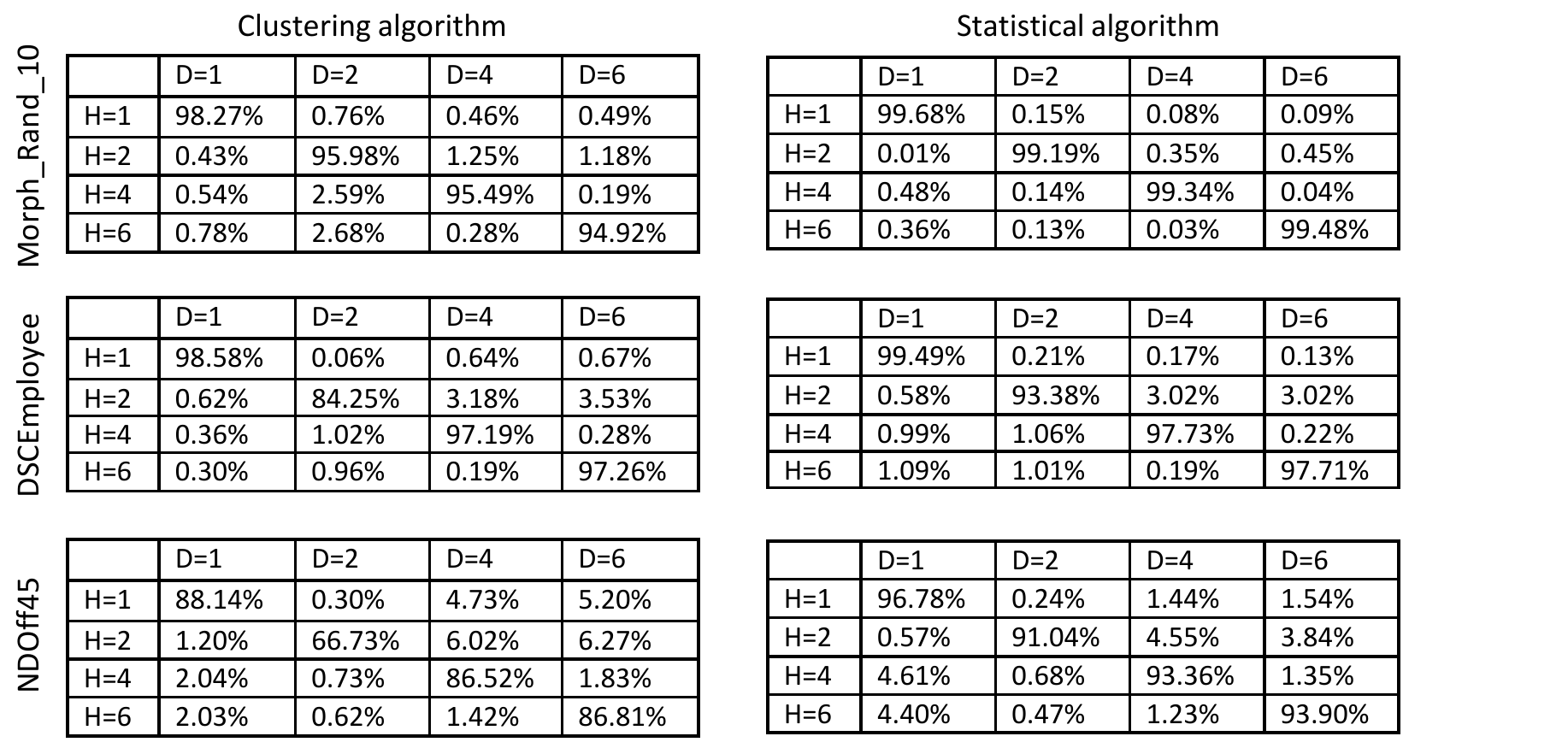}
\par\end{centering}

\protect\caption{Classification error rates for the spectral clustering method (left)
and the statistical method (right) on (3,3) ID pairs from MORPH\_Rand\_10
(top), DSCEmployee (middle) and NDOff45 (bottom).}
\end{figure}

\noindent For each of these cases, we also examine the distributions
of the log-likelihood scores when the classifier reaches the corresponding
decision. The plots are shown in Figure 4. The scores generally look
like Gaussian variables for most of the hypotheses, typically taking
on values between $0$ and $60$ (or $0$ and $4$ after dividing
by the number of edges as described earlier), with the exception of
cases involving $H=2$ or $D=2$. The exact reasons for this are not
well understood, but in the $H=2$, $D=2$ case, this can interpreted
to mean that the classifier is always ``very sure'' of its answer
when it correctly detects a multi-ID case.\\

\noindent We finally consider a different type of test in Figure 5,
where the classifier is run on MORPH\_Rand\_10 for different values
of $M$, with 80,000 ID pairs to be checked. There is no fraud simulation
done here, and the objective is instead to study the false positive
rates across all seven hypotheses and to examine the tradeoff with
the computation time. An ideal classifier would choose $D=1$ for
all 80,000 cases, although in practice, the dataset actually contains
a few mislabeled images that would prevent this. It can be seen that
the $H=3$ and $H=5$ cases generate a lot of false positives relative
to the other cases. There is also zero improvement in going from $M=3$
to $M=4$, despite the large increase in computation time (the time
shown is the total for all 80,000 pairs).

\noindent 
\begin{figure}[H]
\begin{centering}
\includegraphics[trim=2in 0.5in 1in 0.5in, clip=true, scale=0.7]{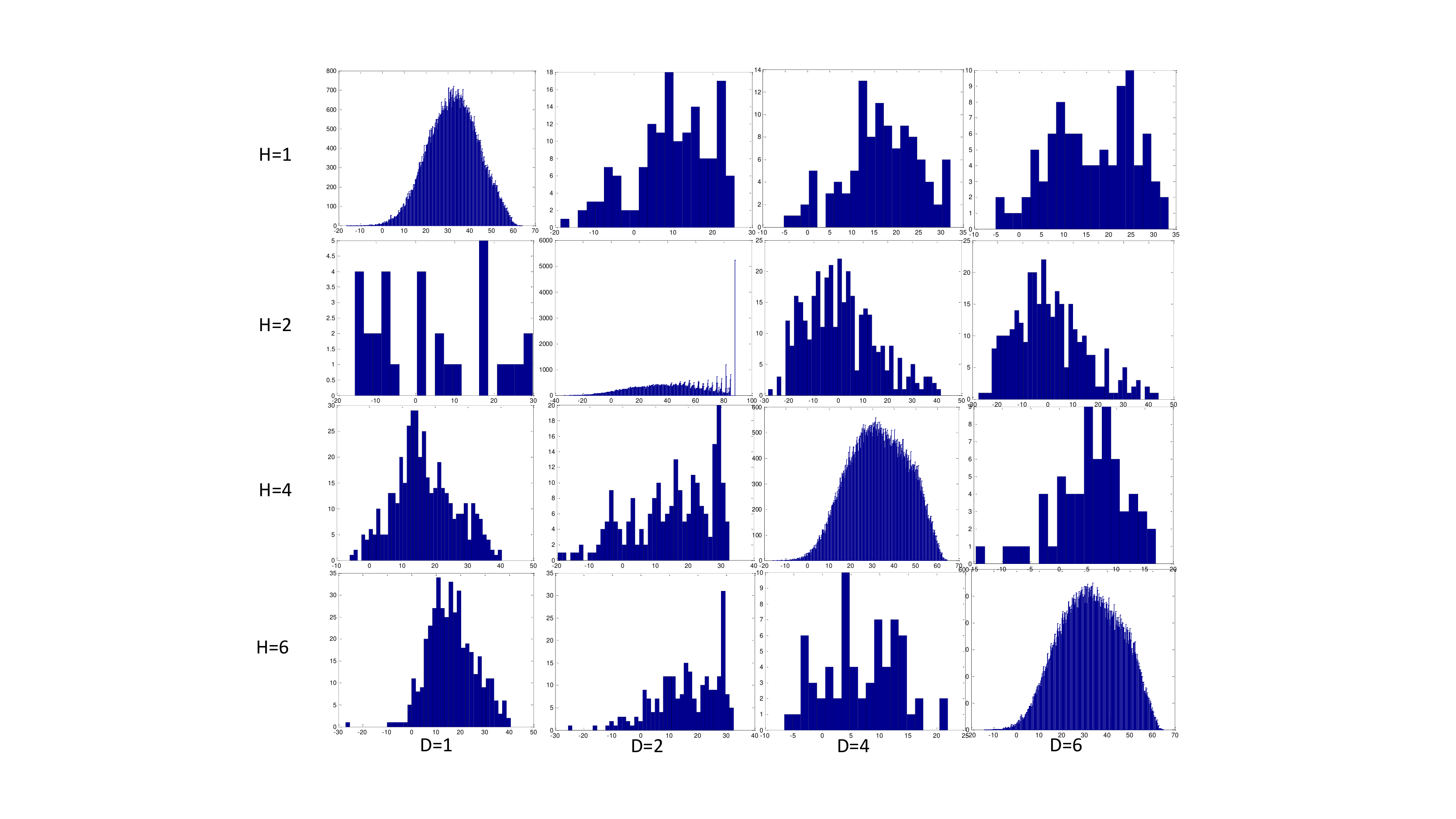}
\par\end{centering}

\protect\caption{Log-likelihood score densities for different hypotheses and decisions.}
\end{figure}

\noindent 
\begin{figure}[H]
\begin{centering}
\includegraphics[trim=3.5in 3.3in 3in 3in, clip=true, scale=1.0]{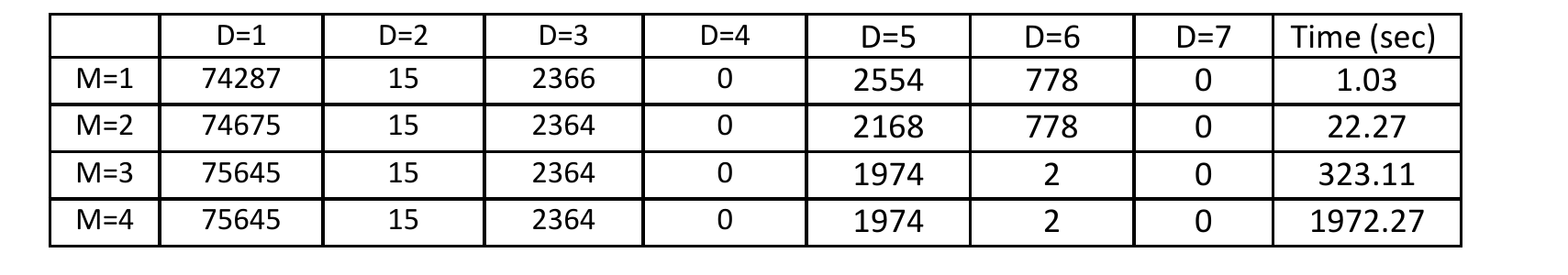}
\par\end{centering}

\protect\caption{Results of the false positive test on 80,000 (10,10) ID pairs from
MORPH\_Rand\_10 for different values of $M$.}
\end{figure}

\bibliographystyle{plain}
\bibliography{Fraud}

\end{document}